\documentclass{article}

\PassOptionsToPackage{numbers}{natbib}

 \usepackage[preprint]{neurips_2026}

\usepackage[utf8]{inputenc} 
\usepackage[T1]{fontenc}    
\usepackage{hyperref}       
\usepackage{url}            
\usepackage{booktabs}       
\usepackage{amsfonts}       
\usepackage{nicefrac}       
\usepackage{microtype}      
\usepackage{xcolor}         
\usepackage{graphicx}
\usepackage{booktabs}
\usepackage[accsupp]{axessibility}  
\usepackage{enumitem} 
\usepackage{multirow} 
\usepackage{colortbl}
\usepackage{hhline}
\usepackage{wrapfig}
\title{SyncDPO: Enhancing Temporal Synchronization in Video-Audio Joint Generation via Preference Learning}

\author{
Xin Cheng$^{1}$\thanks{Equal contribution. Contact: chengxin000@ruc.edu.cn}
\quad
Xihua Wang$^{1}$\footnotemark[1]
\quad
Ying Ba$^{1}$
\quad
Yuyue Wang$^{1}$ \\
\quad
\textbf{Kaisi Guan}$^{1}$ 
\quad
\textbf{Yinbo Wang}$^{1}$
\quad
\textbf{Wenpu Li}$^{2}$
\quad
\textbf{Ruihua Song}$^{1}$\thanks{Corresponding author.} \\
$^{1}$ Renmin University of China
\qquad
$^{2}$ Westlake University
}

\begin{document}
\maketitle

\begin{abstract}
    Recent advancements in video-audio (V-A) joint generation have achieved remarkable success in semantic correspondence. However, achieving precise temporal synchronization, which requires fine-grained alignment between audio events and their visual triggers, remains a challenging problem. The post-training method for audio-visual joint generation is largely dominated by Supervised Fine-Tuning (SFT), but the commonly used Mean Squared Error (MSE) loss provides insufficient penalties for subtle temporal misalignments. Direct Preference Optimization (DPO) offers an alternative by introducing explicit misaligned counterparts to better improve temporal sensitivity. In this paper we propose a post-training framework \textbf{\textsc{SyncDPO}}, leveraging DPO to improve the temporal sensitivity of V-A joint generation. Conventional DPO pipelines typically depend on costly sampling-and-ranking procedures to construct preference pairs, resulting in substantial computational overhead. To improve efficiency, we introduce a suite of on-the-fly, rule-based negative construction strategies that distort temporal structures without incurring additional annotation or sampling overhead. We demonstrate that the temporal alignment capability can be effectively reinforced by providing explicit negative supervision through temporally distorted V-A pairs. Accordingly, we implement a curriculum learning strategy that progressively increases the difficulty of negative samples, transitioning from coarse misalignment to subtle inconsistencies. Extensive objective and subjective experiment results across four diverse benchmarks, ranging from ambient sound videos to human speech videos, demonstrate that SyncDPO significantly outperforms other methods in improving model's temporal alignment capability. It also demonstrates superior generalization on out-of-distribution benchmark by capturing intrinsic motion-sound dynamics. Demo and code is available in \url{https://syncdpo.github.io/syncdpo/}.
\end{abstract}

\begin{figure}[h!]
    \centering
    \includegraphics[width=1\linewidth]{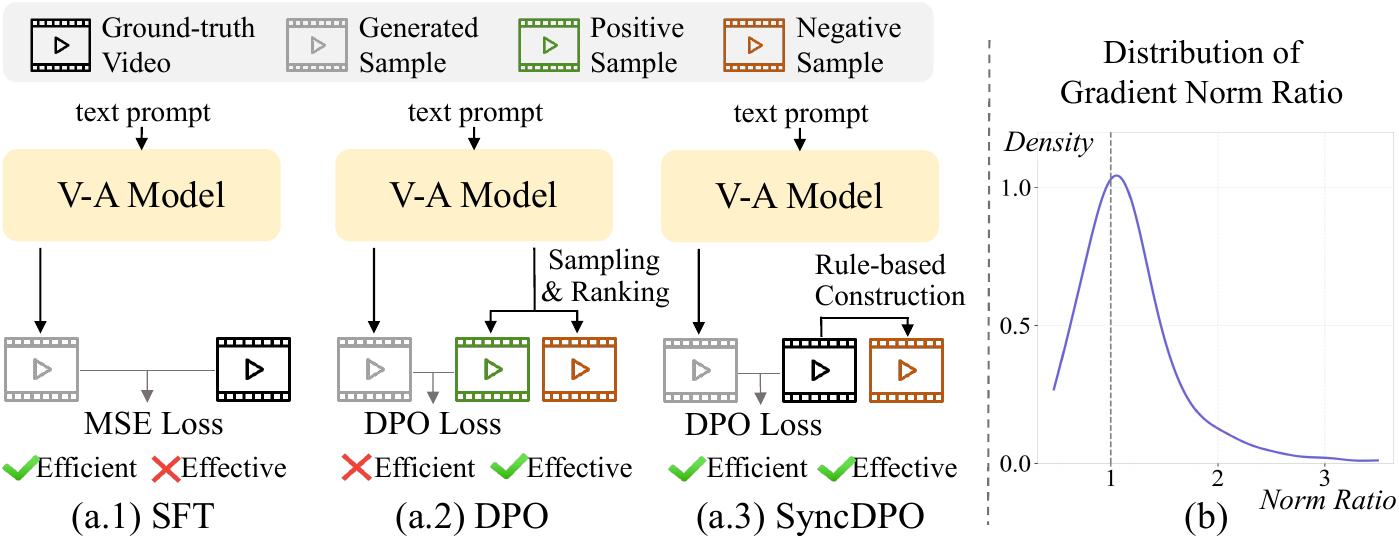}
    
    \caption{A direct comparison of SFT, DPO, and SyncDPO paradigms. 
    (a) shows the difference between different training strategies.
    SFT relies on MSE loss and provides weak supervision for temporal alignment, leading to limited convergence speed. Conventional DPO requires costly sampling-and-ranking to construct preference pairs, resulting in low efficiency. In contrast, SyncDPO achieves both high performance and efficiency by generating temporally misaligned negatives through rule-based perturbations. 
    (b) shows the distribution of gradient norm ratio between unweighted preference score and MSE loss, computed as
    $\|\nabla_\theta L_{\mathrm{SyncDPO}}\|_2 /
    \|\nabla_\theta L_{\mathrm{SFT}}\|_2$, where the SyncDPO gradients are computed from the preference score difference before applying
    $\log\sigma(\cdot)$ and $\beta$. 
    }
    
\label{fig:teaser}
\end{figure}

\section{Introduction}
\label{sec:intro}

Recent advancements in generative AI have significantly improved joint video-audio synthesis. Landmark industrial models such as Veo~\cite{googleveo32025google}, Sora~\cite{sora2}, and Seedance~\cite{seedance2025seedance} have demonstrated outstanding capacity for high-fidelity joint generation, producing visually stunning and acoustically rich contents with strong coherence. In parallel, the academic community has also developed sophisticated frameworks~\cite{low2025ovi, wang2025universe, liu2025javisdit, zhao2025uniform, wang2024avdit, guan2025taming} that bridge powerful pre-trained unimodal generative branches~\cite{wan2025wan,genmo2024mochi,cheng2025MMAudio,cheng2024lova,cheng2025vssflow} for seamless joint synthesis. These approaches leverage massive single-modal data and pretrained models~\cite{wan2025wan, cheng2025MMAudio} to achieve high semantic fidelity and cross-modal alignment.

Although these models can generate high-quality visual content and contextually relevant audio, precise temporal synchronization between the two modalities remains a significant but underexplored challenge~\cite{liu2026javisdit++, huang2025jova}. Current open-source models still sometimes fail to align audio and events accurately in time. 
This issue is particularly evident in timing-sensitive scenarios, such as human speech videos and event-triggered ambient sounds. For instance, gunshot sounds may drift from gunfire animations, or human speech may mismatch lip movements (as shown in Fig.~\ref{fig:case}), which severely degrades perceived realism and immersion.

Currently, the exploration of post-training methods for V-A joint generation model remains relatively limited. 
Supervised Fine-Tuning (SFT) on high-quality data is a common training strategy~\cite{wang2025animate, low2025ovi} to mitigate temporal misalignment. However, its effectiveness is fundamentally limited: using MSE loss as training objective, it excels at overall feature reconstruction but may not provide sufficient guidance for distinguishing precise temporal offsets. 
To address these limitations, we analyze the efficacy of Direct Preference Optimization (DPO)~\cite{rafailov2023direct, liu2025flowdpo} and propose \textbf{\textsc{SyncDPO}} to improve temporal consistency in the V-A post-training process.
Unlike SFT which learns solely from positive examples, DPO-based approach explicitly contrasts synchronized pairs against temporally distorted ones. This discriminative supervision compels the model to recognize and penalize synchronization errors, substantially sharpening temporal sensitivity.
We visualize the distribution of the gradient norm ratios between two different training objectives in Fig.~\ref{fig:teaser}(b). 
The ratio is generally larger than one, indicating that the model receives stronger discrepancy signals by introducing negative examples in the preference learning process, enabling more effective optimization for temporal synchronization.

However, applying DPO to audio-visual models introduces two key challenges: how to construct preference pairs in a cost-efficient manner, and which construction strategies are most effective for improving temporal alignment. Existing DPO pipelines typically rely on repeated sampling, quality ranking, or human annotation to obtain preference data~\cite{liu2025flowdpo, liu2026javisdit++}, bringing substantial computational and labeling overhead. Moreover, they provide insufficient insight into how different preference construction strategies influence temporal alignment.
In our SyncDPO, we systematically investigate different construction strategies and overcome the limitation with an on-the-fly rule-based negative construction method tailored to temporal alignment. During data loading, we directly disrupt the temporal structure of existing synchronized V-A pairs to synthesize negatives, eliminating the need for external labeling or additional sampling. 
We explore several perturbation strategies to investigate their effects on temporal alignment as illustrated in Fig.~\ref{fig:construct}. 
Furthermore, we adopt a curriculum learning strategy~\cite{bengio2009curriculum, curriculumdpo} that progressively increases the training difficulty. 
The model first learns from coarse, easily detectable misalignments to establish basic temporal awareness, then gradually shifts emphasis to more subtle deviations. 
Evaluation results demonstrate that the curriculum strategy is more effective than using a single perturbation type or a uniform mixture, leading to improved synchronization performance.

Extensive validation across four benchmarks covering environmental sound and human speech videos under various training settings demonstrates the superiority of SyncDPO.
Crucially, our approach exhibits superior out-of-domain generalization by capturing intrinsic motion-sound dynamics. 

In summary, our main contributions are:
\textbf{(1)} We propose SyncDPO, leveraging negative supervision to enhance temporal alignment in V-A generation models, which demonstrates clear superiority in achieving precise synchronization.
\textbf{(2)} We design efficient temporal perturbation methods for on-the-fly construction of negative samples, enabling cost-effective supervision without additional overhead. We further introduce a curriculum learning method that progressively refines temporal sensitivity for better performance.
\textbf{(3)} Extensive objective and subjective evaluations on multiple benchmarks confirm SyncDPO's consistent synchronization gains and robust generalization capabilities.

\section{Related work}
\label{sec:related}

\paragraph{Video-audio joint generation.}
Recent video-audio generation methods have significantly advanced joint synthesis of visual and acoustic content.
Early approaches mainly adopt unified single-backbone architectures to generate both modalities within a shared generative framework~\cite{ruan2023mmdiffusion,tang2023codi,wang2024avdit,zhao2025uniform}.
More recent methods move toward dual-tower paradigms, where strong video and audio backbones are connected through cross-modal interaction modules~\cite{liu2025javisdit,wang2025animate,wang2025universe,low2025ovi,huang2025jova,hacohen2026ltx}.
For example, Ovi~\cite{low2025ovi} exchange information between video and audio towers through symmetric cross-attention, while JoVA~\cite{huang2025jova} and JointDiT~\cite{wang2025animate} perform joint modeling over combined video-audio token sequences.
Although they achieve strong semantic correspondence, precise temporal synchronization remains challenging, especially in scenarios like lip-speech motion and event-triggered sounds.
Most prior efforts focus on architecture design and cross-modal fusion, whereas post-training strategies for improving fine-grained temporal alignment remain relatively underexplored.
Our work introduces a preference-learning post-training framework for sharpening temporal precision in V-A generation.

\paragraph{Preference optimization for generation.}
Direct Preference Optimization (DPO)~\cite{rafailov2023direct} has been widely used to align generative models with human preferences without explicitly training a reward model.
In image generation, methods such as DDPO~\cite{black2023ddpo}, DPOK~\cite{fan2023dpok}, Diffusion-DPO~\cite{wallace2024diffusiondpo}, D3PO~\cite{yang2024d3po}, SPIN~\cite{yuan2024self}, and SPO~\cite{liang2025spo} adapt preference optimization to diffusion models through reinforcement learning and preference data.
Preference optimization has also been extended to video generation and flow-matching model, such as VideoDPO~\cite{liu2025videodpo}, OnlineVPO~\cite{zhang2024onlinevpo} and FlowDPO~\cite{liu2025flowdpo}. 
More recently, these DPO pipelines typically require human annotations, external reward models, repeated sampling, or generated candidate ranking to obtain preference data.
Although effective, such procedures introduce substantial computational overhead. In contrast, SyncDPO constructs temporally misaligned negatives on the fly through lightweight rule-based perturbations, avoiding additional sampling or annotation while directly targeting audio-visual synchronization.

\section{Method}
\label{sec:method}
\subsection{Preliminaries}
\label{sec:preliminary}

\paragraph{Flow-matching framework}

Flow Matching~\cite{lipman2022flow} is a generative modeling paradigm that learns a velocity field to transport samples from a simple source distribution $\mathcal{P}(x_0)$ (typically standard Gaussian noise $x_0 \sim \mathcal{N}(0,\mathbf{I})$) to the complex target data distribution $\mathcal{P}(x_1)$ along straight-line paths in continuous time. The dynamics are governed by the following ODE:
\begin{equation}
    x_t = (1-t) x_0 + t x_1, \quad t \sim \mathcal{U}[0,1].
\end{equation}

In the V-A joint generation model, $x_1$ stands for the latent representations of video and audio from pre-trained VAEs.
The training objective of flow matching model is to minimize the discrepancy between the predicted velocity and the ground-truth velocity
\begin{equation}
    \mathcal{L}_{\text{FM}}(\theta) = \mathbb{E}_{t, x_0, x_1} \left[ \| v_\theta(x_t, t) - (x_1 - x_0) \|^2 \right].
\end{equation}

\paragraph{Direct preference optimization}

Direct Preference Optimization (DPO)~\cite{rafailov2023direct} provides an efficient way to align generative models with human preferences, bypassing the need for an explicit reward function. 
Given a prompt $y$ and a preference pair $(x^{w}, x^{l})$, where $x^{w}$ and $x^{l}$ denote the preferred (winner) and dispreferred (loser) samples, respectively, the DPO objective can be expressed in closed form as:
\begin{equation}
    \mathcal{L}_{\text{DPO}}(\pi_\theta; \pi_{\text{ref}}) = -\mathbb{E}_{(y,x^w,x^l) \sim \mathcal{D}} \left[ \log \sigma \left( \beta \log \frac{\pi_\theta(x^w|y)}{\pi_{\text{ref}}(x^w|y)} - \beta \log \frac{\pi_\theta(x^l|y)}{\pi_{\text{ref}}(x^l|y)} \right) \right],
    \label{eq:dpo}
\end{equation}
where $\pi_\theta$ denotes the policy, $\pi_{\text{ref}}$ is the frozen reference model, $\beta > 0$ is a hyperparameter controlling the strength of KL regularization, and $\sigma(\cdot)$ denotes the sigmoid function. 

In flow-matching models, the policy is parameterized by a velocity field $v_\theta(x_t, t)$ that predicts the direction from noisy state $x_t$ to the target data. Following the implementation of FlowDPO~\cite{liu2025flowdpo}, we train our flow-matching model using the subsequent DPO objective:
\begin{equation}
\begin{split}
\mathcal{L}_{\text{Flow-DPO}}(\theta) = -\mathbb{E} \Bigl[ - \log \sigma \Bigl( \beta \bigl( 
& \|v^w - v_\theta(x_t^w, t)\|^2 - \|v^w - v_{\text{ref}}(x_t^w, t)\|^2 \\
& - (\|v^l - v_\theta(x_t^l, t)\|^2 - \|v^l - v_{\text{ref}}(x_t^l, t)\|^2) \bigr) \Bigr) \Bigr],
\end{split}
\label{eq:flow_dpo}
\end{equation}

Intuitively, minimizing $\mathcal{L}_{\text{Flow-DPO}}$ pulls the predicted velocity $v_\theta$ closer to the preferred target's velocity $v^w$ (encouraging temporal synchronization) while pushing it away from the dispreferred velocity $v^l$ (penalizing misalignment), with the reference velocity $v_{\text{ref}}$ providing regularization. The preference signal is driven by differences in prediction errors relative to the reference model.

\subsection{Negative construction method}
\label{sec:construction}

Effective application of DPO in V-A generation critically depends on the quality of preference pairs.
Conventional DPO pipelines usually generate multiple candidates per prompt and then obtain preferences through human annotation, reward-model scoring, or automated metrics, leading to substantial computational and labeling costs.

\begin{figure}[t!]
    \centering

    \includegraphics[width=1\linewidth]{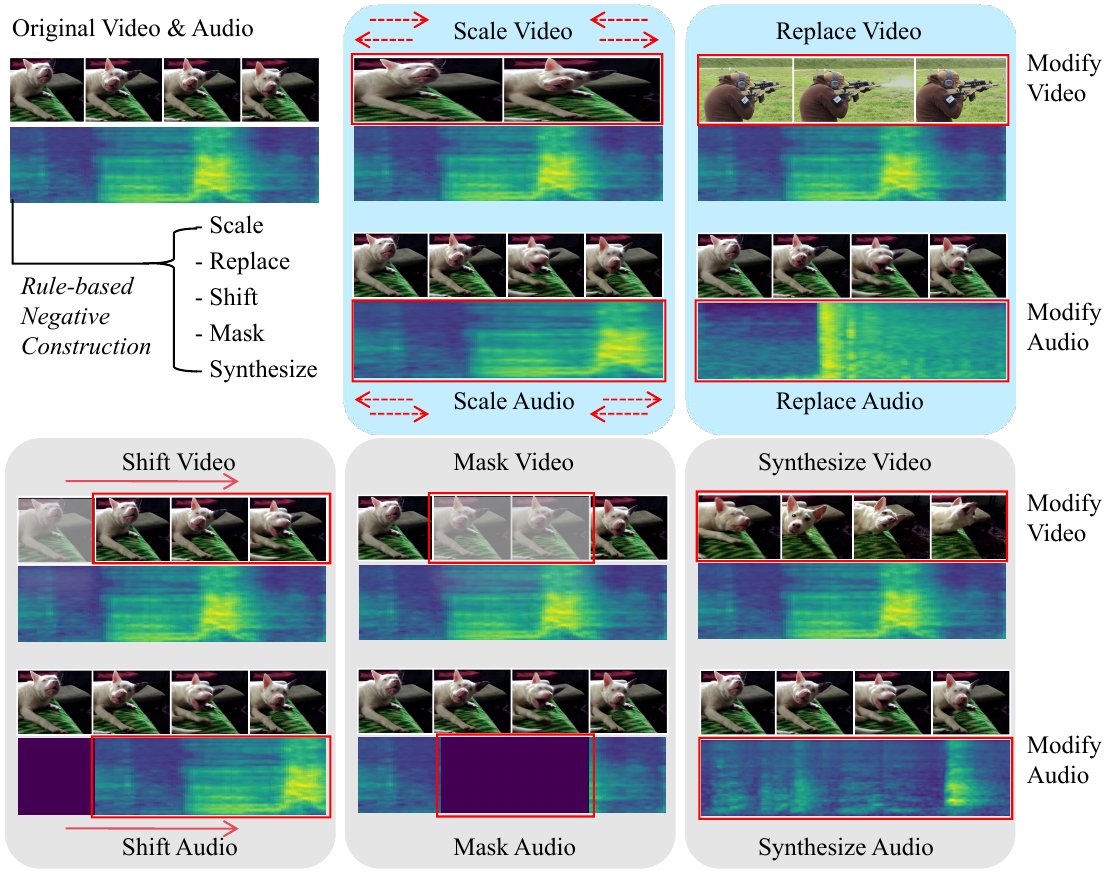}
    
    \caption{Visualization of on-the-fly negative construction for SyncDPO. The video or audio modality in the original synchronized pair (top left) is temporally perturbed using five rule-based operations: Scaling, Replacing, Shifting, Masking, and Synthesizing. 
    }
    \label{fig:construct}
    
\end{figure}

In SyncDPO, we adopt a lightweight on-the-fly negative construction strategy.
Instead of additional model sampling or preference collection, we derive temporally misaligned negatives directly from synchronized training pairs during data loading.
For each positive pair $(x^w, y)$, where $x^w$ denotes the synchronized video-audio sequence and $y$ is the corresponding text prompt, we construct a negative counterpart $(x^l, y)$ by perturbing the ground-truth aligned data.
We consider five perturbation operations that simulate common audio-visual misalignment patterns as in Fig.~\ref{fig:construct}:

\begin{itemize}
    \item \textbf{Scaling}: Rescale the temporal dimension of either video or audio by a scale factor $s \sim \mathcal{U}[0.5, 1.5]$, creating speed inconsistency between modalities.
    
    \item \textbf{Replacing}: Substitute the video or audio sequence with another sample from the dataset, introducing strong semantic and temporal discord.
    
    \item \textbf{Shifting}: Apply a temporal offset $\delta \sim \mathcal{U}[-2.5\,\text{s}, 2.5\,\text{s}]$ to video or audio sequence, simulating advance/delay synchronization errors.
    
    \item \textbf{Masking}: Mask either the video or audio sequence with a mask ratio $m \sim \mathcal{U}[0.1, 0.3]$, creating partially silent audio or frozen visual content.
    
    \item \textbf{Synthesizing}: Replace video or audio with a generated sample from the no-tuned reference policy $\pi_{\text{ref}}$, producing realistic but imperfect negatives.
\end{itemize}

Each perturbation is applied to either video or audio with equal probability, so the negative sample partially preserves prompt semantics while violating temporal correspondence.
Preliminary experiments show that \textbf{Replacing} and \textbf{Scaling} provide particularly effective supervision for both human-speech and ambient-sound benchmarks. Detailed results are provided in Sec.~\ref{sec:exp_on_construct}.

Compared to conventional DPO pipeline, this construction is efficient and scalable: it requires no extra sampling, pre-collected preference data, or human annotation, while producing diverse and discriminative negatives with clear temporal disruptions.

\subsection{Curriculum learning}
\label{sec:curriculum}

DPO training is sensitive to the difficulty distribution of negative examples.
Overly hard negatives may hinder early optimization before the model establishes basic temporal discrimination, while overly easy negatives provide limited supervision for fine-grained synchronization~\cite{curriculumdpo}.
Motivated by curriculum learning~\cite{bengio2009curriculum,curriculumdpo}, we adopt a difficulty-scheduled strategy over Replacing and Scaling, the two most effective perturbations identified in Sec.~\ref{sec:exp_on_construct}.

Specifically, we dynamically adjust the sampling probabilities of Replacing and Scaling. Replacing produces negatives with both semantic and temporal mismatches, making them easier to distinguish. While Scaling preserves semantic consistency but introduces temporal distortions, yielding harder negatives that require finer temporal discrimination. We start from uniform sampling and gradually shift the sampling probability from Replacing to Scaling during training. 
Formally, let $p_{\text{replace}}(t)$ and $p_{\text{scale}}(t)$ denote their sampling probabilities at training step $t$. We schedule them as:
\begin{equation}
\begin{aligned}
p_{\text{replace}}(t) &= 0.5 - kt, \\
p_{\text{scale}}(t) &= 1 - p_{\text{replace}}(t).
\end{aligned}
\end{equation}
where $k$ controls the transition rate from Replaced to Scaled negatives.
Other perturbation types, like Shifting, Masking, and Synthesizing, are excluded from the curriculum strategy.
This progressive schedule guides the model to learn from coarse semantic-temporal discrimination to fine-grained temporal alignment, and its effectiveness is validated in Sec.~\ref{sec:exp_curriculum}.

\section{Experiment}
\label{sec:exp}
\subsection{Experiment setup}

\paragraph{Benchmarks and datasets.}
\label{sec:dataset}
We evaluate SyncDPO on four benchmarks covering both human-speech and ambient-sound videos: LRS2~\cite{afouras2018LRS2}, AVSync~\cite{zhang2024avsync}, GreatestHits~\cite{owens2016greatesthits}, and VABench~\cite{hua2025vabench}.
LRS2 is used for lip-speech synchronization evaluation, while AVSync and GreatestHits focus on event-level audio-visual synchronization. 
VABench serves as the out-of-domain benchmark for assessing generalization to more diverse and challenging synchronization scenarios.
For training data, we consider both \textit{in-domain} and \textit{out-of-domain} settings to validate the robustness of our proposed method.
The in-domain set contains 5,000 synchronized video-audio pairs from LRS2, AVSync, and GreatestHits, while the out-of-domain set contains 10,000 clips sampled from Koala~\cite{wang2025koala}.
Detailed dataset statistics and preprocessing procedures are provided in Appendix~\ref{appendix:datasets}.

\paragraph{Evaluation metrics.}
\label{sec:eval}
We evaluate both synchronization accuracy and generation quality for human-speech and ambient-sound videos.
For human-speech videos, we use LSE-D and LSE-C~\cite{lse} as the primary lip-synchronization metrics
We also report WER, MCD, and UTMOS to assess speech intelligibility, acoustic distortion, and perceptual speech naturalness.
For ambient-sound videos, we use DeSync~\cite{iashin2024synchformer,cheng2025MMAudio} as the primary temporal synchronization metric and VA-IB~\cite{girdhar2023imagebind} to measure video-audio semantic consistency.
In addition, we report FAD for audio quality, FVD and KVD for video quality, and CLAP/CLIP scores for text-audio and text-video alignment.
The auxiliary metrics serve as controls to verify that improved synchronization does not come at the cost of generation fidelity or semantic alignment.
Detailed information is provided in Appendix~\ref{appendix:metrics}.

\paragraph{Implementation details.}
\label{sec:detail}
Our model is built upon the Ovi framework~\cite{low2025ovi}, which has demonstrated superior performance in high-fidelity joint video-audio generation. 
All training samples are truncated to 5 seconds with audio resampled to 16 kHz, following the default setting of Ovi. 
We implement SFT, vanilla DPO~\cite{liu2025flowdpo}, and SyncDPO for training.
Following prior work VA-DPO~\cite{liu2026javisdit++}, we construct preference pairs for vanilla DPO by generating and ranking multiple candidates. Specifically, for each text prompt, we sample three videos and evaluate their temporal alignment using DeSync~\cite{cheng2025MMAudio}. The sample with the lowest DeSync score is selected as the preferred sample, while the one with the highest DeSync score is treated as the dispreferred sample. 
Under both in-domain and out-of-domain training settings, DPO and SyncDPO are trained for one epoch with a batch size of 1, while SFT is trained for two epochs with a batch size of 2, ensuring a comparable number of training steps and training time across methods.
More training details can be found in Appendix~\ref{appendix:tab_hyperparams}.

\subsection{Experiments on different negative construction methods}
\label{sec:exp_on_construct}
We first conduct experiments to investigate the effectiveness of various negative construction strategies described in Sec.~\ref{sec:construction}. 
The evaluation results on LRS2 and AVSync15 benchmarks are summarized in Tab.~\ref{tab:cons}. 
More training details and evaluation results on other benchmarks can is in Appendix~\ref{appendix:exp_construction}.

\begin{table}[t!]
\centering
\caption{
Evaluation results of different negative construction methods on AVSync15 and LRS2 benchmarks. We focus on synchronization metrics: LSE-D, LSE-C and DeSync, which are blue-highlighted. 
\textbf{Bold} and \underline{underline} indicate the best and second-best performance, respectively. 
}
\label{tab:cons}

\renewcommand{\arraystretch}{1.05}
\setlength{\tabcolsep}{2pt}

\resizebox{\textwidth}{!}{
\begin{tabular}{
l
>{\columncolor{blue!10}}c
>{\columncolor{blue!10}}c
ccc
>{\columncolor{blue!10}}c
c
cc
}
\toprule[1.2pt]

& \multicolumn{5}{c}{\textbf{LRS2}} 
& \multicolumn{4}{c}{\textbf{AVSync15}} \\

\cmidrule(lr){2-6} \cmidrule(lr){7-10}

Method
& \multicolumn{2}{c}{Lip Alignment}
& \multicolumn{3}{c}{Speech Quality}
& \multicolumn{2}{c}{VA Alignment}
& \multicolumn{2}{c}{VA Quality} \\

\cmidrule(lr){2-3} \cmidrule(lr){4-6} \cmidrule(lr){7-8} \cmidrule(lr){9-10}

& \cellcolor{blue!10}LSE-D $\downarrow$
& \cellcolor{blue!10}LSE-C $\uparrow$
& WER $\downarrow$ 
& MCD $\downarrow$ 
& UTMOS $\uparrow$
& \cellcolor{blue!10}DeSync $\downarrow$
& VA-IB $\uparrow$
& FVD $\downarrow$
& $\text{FAD}_{pan}$ $\downarrow$ \\

\midrule[1.2pt]

No tune
& 8.15 & 7.08 & 15.40 & 19.06 & 3.34
& 0.80 & 0.33 & 388.31 & 43.31 \\

\cmidrule(lr){1-10}

Scale
& \underline{7.96} & \textbf{7.12} & 15.70 & 18.88 & 3.39
& \textbf{0.49} & 0.33 & 372.07 & 42.11 \\

Replace
& \textbf{7.92} & \underline{7.07} & 15.10 & 18.50 & 3.38
& \underline{0.54} & 0.33 & 388.15 & 41.59 \\

Shift
& 8.33 & 6.57 & 15.90 & 18.65 & 3.31
& \underline{0.54} & 0.33 & 382.84 & 41.18 \\

Mask
& 8.61 & 6.60 & 16.20 & 18.52 & 3.22
& 0.93 & 0.25 & 1748.33 & 45.59 \\

Synthe.
& 8.04 & 6.80 & 18.40 & 18.95 & 3.28
& \textbf{0.49} & 0.33 & 473.27 & 40.20 \\

\bottomrule[1.2pt]
\end{tabular}
}
\end{table}

As shown in Tab.~\ref{tab:cons}, \textbf{Replacing} and \textbf{Scaling} achieve the strongest overall synchronization performance across both benchmarks.
Replacing provides clear semantic-temporal contrast by constructing negatives that differ from positives in both content and timing, making it especially effective for establishing coarse alignment boundaries.
Scaling preserves semantic consistency while distorting temporal structure, producing harder negatives that encourage fine-grained temporal discrimination.
These complementary properties motivate their use in our curriculum schedule.

Other perturbations are less favorable.
Shifting improves ambient-sound synchronization by introducing event-level temporal offsets, but is less effective for human speech, where global shifts may not match typical lip-speech misalignment patterns.
Synthesizing is competitive on AVSync15 but requires additional sampling from the reference policy, increasing training cost.
Masking performs worst, as its destructive perturbations severely disrupt semantic structure and degrade generation quality.
Overall, Replacing and Scaling improve synchronization while maintaining stable quality and alignment metrics, indicating that effective temporal negatives can enhance synchronization sensitivity without sacrificing generation fidelity.

\begin{table}[t!]
    \centering
    
    \caption{
        Evaluation results on multiple human speech and ambient sound benchmarks.
        Training Data denotes the datasets utilized for SFT, DPO and SyncDPO training. The term ``ID'' refers to the In-Domain data including AVSync, GreatestHits and LRS2, and  ``OOD'' refers to the Out-of-Domain dataset Koala. The blue-highlighted metrics are synchronization indicators which we primarily focus on.  Since VABench provides only text prompts, generation quality metrics are not evaluated on this benchmark. \textbf{Bold} denote the best performance.
    }
    \label{tab:results_speech}

    \setlength{\tabcolsep}{2.0pt}  
    \renewcommand{\arraystretch}{1.07} 
    \resizebox{\textwidth}{!}{
        \begin{tabular}{l l l >{\columncolor{blue!8}[0.25pt][0.25pt]}c>{\columncolor{blue!8}[0.25pt][0.25pt]}c ccc ccc}
            \toprule[1.4pt]
            
            \multicolumn{11}{c}{\textbf{Human Speech Benchmark}} \\
            \midrule[1.4pt]

            \multirow{2}{*}{Benchmark} & 
            \multirow{2}{*}{Data} & 
            \multirow{2}{*}{Method} & 
            \multicolumn{2}{c}{Lip Alignment} & 
            \multicolumn{3}{c}{Speech Quality} & 
            \multicolumn{3}{c}{Video Quality} \\

            \cmidrule(lr){4-5}
            \cmidrule(lr){6-8}
            \cmidrule(lr){9-11}
            
            & & & LSE-D $\downarrow$ & LSE-C $\uparrow$ & WER $\downarrow$ & MCD $\downarrow$ & UTMOS $\uparrow$ & FVD $\downarrow$ & KVD $\downarrow$ & CLIP $\uparrow$ \\
            \midrule[1.4pt]
    
            \multirow{7}{*}{LRS2}
            & - & No Tune 
            & 8.15 & 7.08 & 15.40 & 19.06 & 3.34 & 250.22 & 0.44 & 0.21 \\
            \cmidrule(lr){2-11}
    
            & \multirow{3}{*}{\shortstack[l]{ID}}
            & SFT 
            & 8.39 & 6.75 & 14.76 & 18.98 & 3.28 & 245.74 & 0.44 & 0.21 \\
            & 
            & DPO 
            & 8.19 & 7.05 & 14.60 & 19.18 & 3.33 & 262.12 & 0.45 & 0.21 \\
            & 
            & SyncDPO 
            & \textbf{7.98} & \textbf{7.18} & 15.20 & 18.68 & 3.40 & 238.98 & 0.43 & 0.21 \\
            \cmidrule(lr){2-11}
    
            & \multirow{3}{*}{\shortstack[l]{OOD}}
            & SFT 
            & 8.20 & 6.90 & 15.28 & 19.40 & 3.31 & 240.55 & 0.43 & 0.21 \\
            & 
            & DPO 
            & 8.19 & 7.04 & 15.36 & 18.92 & 3.30 & 262.67 & 0.46 & 0.21 \\
            & 
            & SyncDPO 
            & \textbf{7.83} & \textbf{7.18} & 15.10 & 18.90 & 3.38 & 240.85 & 0.43 & 0.21 \\
    
            \bottomrule[1.4pt]
        \end{tabular}
    }

    \vspace{0.2em}
    \setlength{\tabcolsep}{2.0pt}  
    \renewcommand{\arraystretch}{1.07} 
    \resizebox{\textwidth}{!}{
        \begin{tabular}{l c l >{\columncolor{blue!8}[0.25pt][0.25pt]}cc cc cc cc}
            \multicolumn{11}{c}{\textbf{Ambient Sound Benchmarks}} \\
            \midrule[1.4pt]

            \multirow{2}{*}{Benchmark} &
            \multirow{2}{*}{Data} & 
            \multirow{2}{*}{Method} & 
            \multicolumn{2}{c}{VA Alignment} &
            \multicolumn{2}{c}{Text Alignment} &
            \multicolumn{2}{c}{Audio Quality} &
            \multicolumn{2}{c}{Video Quality} \\
            \cmidrule(lr){4-5}
            \cmidrule(lr){6-7}
            \cmidrule(lr){8-9}
            \cmidrule(lr){10-11}
            & & 
            & DeSync $\downarrow$ & VA-IB $\uparrow$ 
            & CLAP $\uparrow$ & CLIP $\uparrow$ 
            & $\text{FAD}_{pan}$ $\downarrow$ & $\text{FAD}_{pas}$ $\downarrow$
            & FVD $\downarrow$ & KVD $\downarrow$ \\
            \midrule[1.4pt]
    
            \multirow{7}{*}{AVSync15}
            & - & No Tune
            & 0.81 & 0.33 & 0.48 & 0.26 & 43.31 & 422.88 & 388.31 & 6.03 \\
            \cmidrule(lr){2-11}
    
            & \multirow{3}{*}{\shortstack[l]{ID}}

            & SFT
            & 0.71 & 0.33 & 0.51 & 0.26 & 41.91 & 439.30 & 407.79 & 5.81 \\
            &
            & DPO
            & 0.72 & 0.33 & 0.49 & 0.26 & 43.38 & 421.53 & 412.12 & 6.64 \\
            &
            & SyncDPO
            & \textbf{0.56} & 0.33 & 0.50 & 0.25 & 41.68 & 423.80 & 388.17 & 5.78 \\
            \cmidrule(lr){2-11}
    
            & \multirow{3}{*}{\shortstack[l]{OOD}}
            & SFT
            & 0.92 & 0.34 & 0.49 & 0.25 & 39.19 & 421.54 & 394.60 & 5.93 \\
            &
            & DPO
            & 0.75 & 0.33 & 0.49 & 0.26 & 43.44 & 419.01 & 397.81 & 6.24 \\
            &
            & SyncDPO
            & \textbf{0.67} & 0.34 & 0.49 & 0.25 & 40.49 & 427.92 & 382.99 & 5.76 \\
            \midrule[0.8pt]
    
            \multirow{7}{*}{\shortstack[l]{Greatest \\ Hits}}
            & - & No Tune
            & 0.47 & 0.15 & 0.34 & 0.22 & 48.19 & 506.01 & 352.69 & 1.68 \\
            \cmidrule(lr){2-11}
    
            & \multirow{3}{*}{\shortstack[l]{ID}}
            & SFT
            & 0.43 & 0.14 & 0.34 & 0.22 & 43.33 & 474.38 & 357.53 & 1.45 \\
            &
            & DPO
            & 0.36 & 0.16 & 0.33 & 0.22 & 48.29 & 514.17 & 355.43 & 1.78 \\
            &
            & SyncDPO
            & \textbf{0.30} & 0.15 & 0.34 & 0.22 & 44.30 & 472.97 & 368.62 & 1.63 \\
            \cmidrule(lr){2-11}
    
            & \multirow{3}{*}{\shortstack[l]{OOD}}
            & SFT
            & 0.35 & 0.17 & 0.35 & 0.22 & 44.85 & 469.70 & 322.36 & 1.50 \\
            &
            & DPO
            & 0.38 & 0.16 & 0.34 & 0.22 & 47.47 & 500.18 & 349.50 & 1.69 \\
            &
            & SyncDPO
            & \textbf{0.29} & 0.18 & 0.34 & 0.21 & 41.27 & 433.99 & 383.28 & 1.78 \\
            \midrule[0.8pt]
    
            \multirow{7}{*}{VABench}
            & - & No Tune
            & 0.70 & 0.23 & 0.39 & 0.24 & - & - &  -& - \\
            \cmidrule(lr){2-11}
    
            & \multirow{3}{*}{\shortstack[l]{ID}}

            & SFT
            & 0.54 & 0.20 & 0.39 & 0.24 & - & - & - & - \\
            &
            & DPO
            & 0.67 & 0.23 & 0.40 & 0.24 & - & - & - & - \\
            &
            & SyncDPO
            & \textbf{0.51} & 0.23 & 0.39 & 0.24 & - & - & - &  -\\
            \cmidrule(lr){2-11}
    
            & \multirow{3}{*}{\shortstack[l]{OOD}}
            & SFT
            & 0.49 & 0.22 & 0.39 & 0.24 & - & - & - &  -\\
            &
            & DPO
            & 0.68 & 0.24 & 0.39 & 0.24 & - & - & - &  -\\
            &
            & SyncDPO
            & \textbf{0.43} & 0.25 & 0.39 & 0.23 & - & - & - &  -\\
    
            \bottomrule[1.4pt]
        \end{tabular}
    }
\end{table}

\subsection{Main results}

We compare SyncDPO with No-tune baseline, SFT, and vanilla DPO methods across four benchmarks, as shown in Tab.~\ref{tab:results_speech}.
SyncDPO consistently achieves the best temporal alignment under both in-domain and out-of-domain settings, while maintaining competitive generation quality.

On LRS2 benchmark, SyncDPO better captures the intrinsic correspondence between speech audio and lip motion by explicitly contrasting synchronized and misaligned pairs. 
SyncDPO obtains the best LSE-D and LSE-C, indicating improved lip-speech synchronization.
In contrast, SFT can degrade synchronization compared to the No-tune baseline, suggesting that reconstruction-based objectives are insufficient for fine-grained temporal alignment and may fit the nuisance visual variations such as camera motion, head pose changes, and partial lip occlusions, leading to degraded synchronization.

On ambient-sound benchmarks, SyncDPO consistently achieves the best synchronization performance while maintaining strong semantic consistency. It also demonstrate better generalization than SFT.
On both AVSync15 and GreatestHits benchmarks, SyncDPO achieves significant improvements in synchronization performance compared with both SFT and vanilla DPO.
On the out-of-domain benchmark VABench, although SFT improves DeSync under the in-domain training setting, it causes a noticeable degradation in VA-IB, indicating weakened cross-modal semantic consistency. This suggests that SFT may fit to the training distribution and rely on dataset-specific correlations rather than learning robust temporal alignment. Such overfitting becomes even more pronounced in the low-data regime.
Besides, contrast to vanilla DPO, SyncDPO constructs informative temporal preference pairs, leading to more effective  supervision. 
Overall, these results show that SyncDPO improves temporal synchronization without sacrificing semantic alignment or generation quality.

\subsection{Results on curriculum training}
\label{sec:exp_curriculum}

\begin{wraptable}[14]{r}{0.56\textwidth}
\vspace{-1em}
\centering
\small
\setlength{\tabcolsep}{2.0pt}
\renewcommand{\arraystretch}{1.25}
\caption{Evaluation results on LRS2 and VABench under different negative-sampling schedules.  $k$ denotes the rate of change in the sampling probability in \%.}
\label{tab:ablation_curriculum}
\vspace{0.8em}
\resizebox{\linewidth}{!}{
\begin{tabular}{lcccc}
\toprule
Strategy & LSE-D$\downarrow$ & LSE-C$\uparrow$ & DeSync$\downarrow$ & VA-IB$\uparrow$ \\
\midrule
Scale only 
& \underline{7.86} & 7.15 & \underline{0.57} & 0.22 \\
Replace only 
& 7.87 & 7.12 & 0.58 & \underline{0.24} \\
Uniform ($k=0$) 
& \underline{7.86} & 7.19 & 0.59 & \underline{0.24} \\
Curr. ($k=0.005$) 
& \underline{7.86} & 7.18 & 0.60 & \textbf{0.25} \\
Curr. ($k=0.010$) 
& \textbf{7.83} & \textbf{7.26} & \textbf{0.54} & \textbf{0.25} \\
Curr. ($k=0.015$) 
& 7.87 & \underline{7.21} & 0.59 & 0.23 \\
\bottomrule
\end{tabular}
}
\end{wraptable}

To validate the effectiveness of the proposed curriculum learning strategy, we conduct an ablation study on VABench and LRS2. 
We compare four variants: (1) using only Scaling to construct negative samples throughout training, (2) using only Replacing perturbations, (3) using a uniform mixture (50\% Replacing + 50\% Scaling) at each step, and (4) the proposed curriculum strategy described in Sec.~\ref{sec:curriculum}.
All variants are trained using koala data under identical settings.
Results are summarized in Tab.~\ref{tab:ablation_curriculum}
and more training details can be found Appendix~\ref{appendix:exp_hyperparams}. 

From the results, the curriculum strategy achieves best results by gradually shifting from easier to harder negatives, leading to more stable optimization, improved temporal sensitivity, and better cross-modal semantic alignment. These results validate the effectiveness of difficulty-scheduled curriculum learning in SyncDPO training, consistent with prior findings~\cite{curriculumdpo}.

In contrast, single perturbations or a uniform mixture perform worse on DeSync, LSE-D, and LSE-C, while using only Scaling slightly degrades semantic alignment, as indicated by lower VA-IB scores. 
In addition, the scheduling rate $k$ also plays an important role, as excessively large or small values both lead to suboptimal performance.
The curriculum variant with $k=0.01$ achieves the best overall performance, yielding the best DeSync, VA-IB, LSE-D, and LSE-C scores.
Overly rapid scheduling changes may prevent the model from sufficiently learning coarse alignment patterns, while overly slow transitions can hinder the learning of fine-grained temporal inconsistencies.


    

\subsection{Subjective evaluation and analysis}

\begin{figure}[t!]
    \centering
    \includegraphics[width=1.0\linewidth]{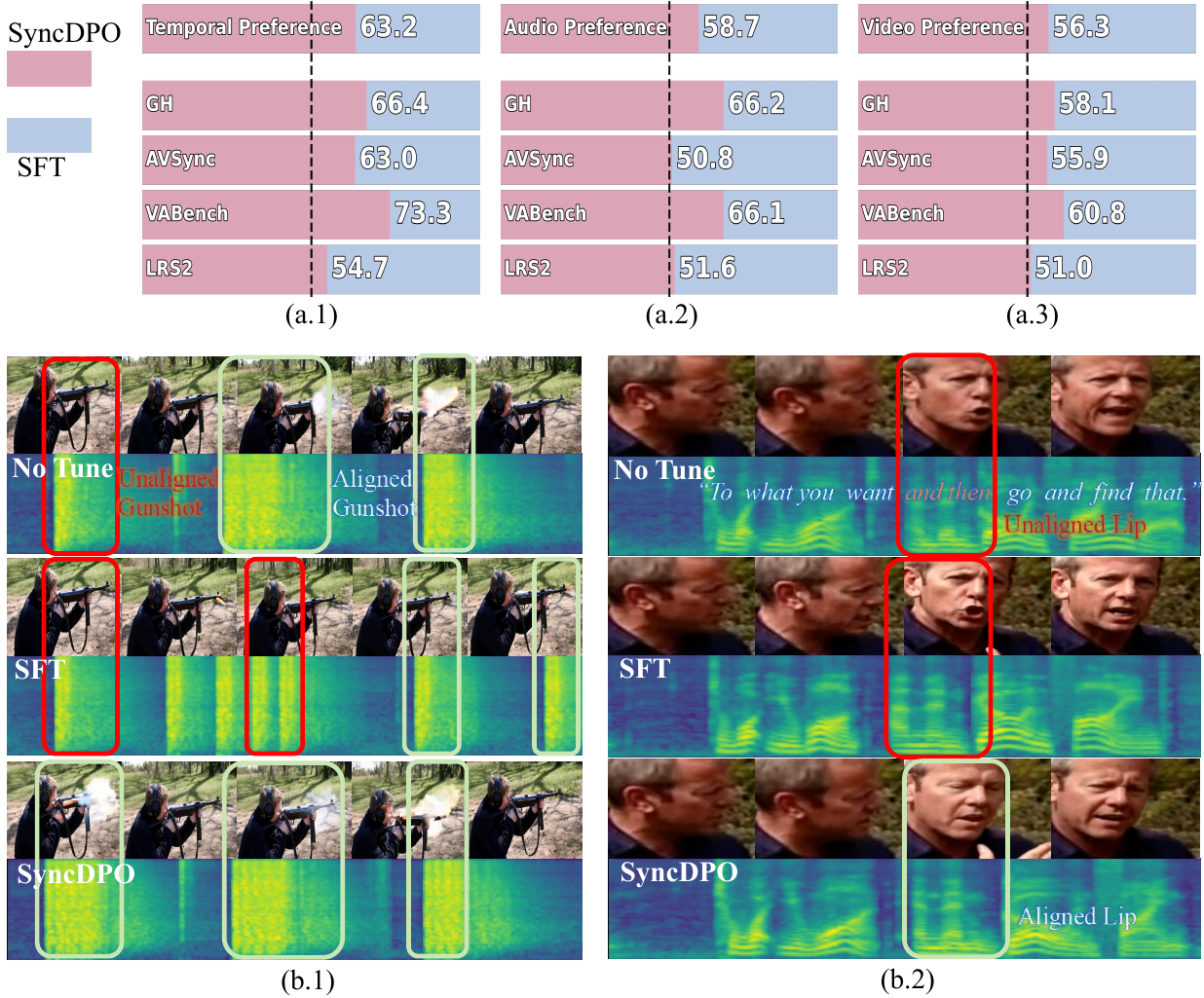}

    \caption{
Human preference evaluation and qualitative comparison between SFT and SyncDPO.
(a.1)--(a.3) report blind pairwise human preferences on temporal alignment, audio quality, and video quality, respectively. The top bar shows the overall preference across all samples, while the remaining bars report per-benchmark results. 
(b) shows qualitative comparisons on event-sound and human-speech examples, respectively. Models trained with SFT still produce audio that is temporally misaligned with the video content, whereas SyncDPO achieves tighter V-A alignment.
    }
    \label{fig:human_eval}
    \label{fig:case}
\end{figure}

To complement automatic metrics, we conduct a blind pairwise human evaluation between SyncDPO and SFT on 40 clips, with 10 samples from each benchmark where the No-tune model shows clear temporal misalignment.
For each clip, outputs from SyncDPO and SFT are presented to 10 qualified annotators, who independently compare them along three aspects: temporal alignment, audio quality, and video quality.
The human evaluation results are in line with the objective metrics. As shown in Fig.~\ref{fig:human_eval}(a), SyncDPO is consistently preferred for temporal alignment while maintaining competitive audio and video quality.
Notably, on the OOD VABench, SyncDPO outperforms SFT across all aspects, further validating its robustness.
Fig.~\ref{fig:case}(b) provides qualitative comparison demos on event-sound and human-speech examples. In (b.1), SFT still produces gunshot audio that is temporally misaligned with the firing moment, while SyncDPO aligns the generated sound with the corresponding visual trigger. In (b.2), SFT shows noticeable lip-audio asynchrony, whereas SyncDPO achieves tighter speech-lip alignment.
We politely invite readers to check the demo page to compare models after training with different methods and experience the advantage of SyncDPO. 

\section{Conclusion}
\label{sec:conc}

We introduce SyncDPO, a DPO-based post-training framework for improving fine-grained temporal synchronization in video-audio joint generative models. By leveraging lightweight rule-based temporal perturbations and a simple curriculum training strategy, SyncDPO addresses key limitations of SFT and conventional DPO while remaining highly efficient. Extensive objective and subjective evaluations across human speech, ambient sound, and out-of-distribution benchmarks demonstrate the strong robustness and effectiveness of our method.


\bibliographystyle{plainnat}
\bibliography{main}

\clearpage
\appendix
\label{sec:appd}

\section{Experiment details}
\label{appendix:exp_details}

\subsection{Benchmarks and training data}
\label{appendix:datasets}

\paragraph{Benchmarks.}
We evaluate on four benchmarks covering both human-speech and ambient-sound videos.
For human-speech evaluation, we use the LRS2~\cite{afouras2018LRS2} test set, where 200 clips are filtered to 3--5 seconds to ensure complete and coherent speech content.
For ambient-sound evaluation, we use AVSync~\cite{zhang2024avsync} and GreatestHits~\cite{owens2016greatesthits}.
AVSync contains high-quality clips curated from VGGSound~\cite{chen2020vggsound}, covering diverse categories such as animals, instruments, and environmental events.
GreatestHits focuses on onset-driven interactions, such as hitting and scratching objects, making it suitable for evaluating event-level synchronization.
We further use the VABench~\cite{hua2025vabench} synchronization test set as an out-of-domain benchmark, which contains 128 prompts designed to assess temporal alignment in more diverse and challenging scenarios.

\paragraph{Training data.}
We consider two training settings: in-domain and out-of-domain.
For the in-domain setting, we curate 5,000 synchronized video-audio pairs, including 1,400 clips from LRS2, 1,300 clips from AVSync, and 2,300 clips from GreatestHits.
For the out-of-domain setting, we randomly sample 10,000 clips from Koala~\cite{wang2025koala}, a large-scale video dataset containing diverse video-audio pairs, including ambient sounds, human speech, and mixed sound-speech scenarios.
Koala contains more unconstrained compositions, off-screen sounds, and complex audio-visual interactions, making it suitable for evaluating generalization.

\paragraph{Preprocessing.}
All training videos are truncated to 5 seconds, and audio is resampled to 16 kHz.
For LRS2, we retain clips with durations between 3 and 5 seconds to ensure coherent speech content.
For speech-containing clips from Koala, we generate transcripts using Whisper~\cite{radford2023whisper} and keep only English utterances with a no-speech probability below 0.1.
For text conditioning, we follow~\cite{guan2025taming} and generate separate descriptive captions for video and audio modalities using Qwen-VL~\cite{Qwen2.5-VL} and Qwen-Omni~\cite{Qwen2.5-Omni}.

\subsection{Evaluation metrics}
\label{appendix:metrics}

\paragraph{Human-speech evaluation.}
For human-speech videos, we evaluate lip-speech synchronization using LSE-D and LSE-C, computed by SyncNet~\cite{lse}.
LSE-D measures the distance between speech and lip embeddings, where a lower value indicates better synchronization.
LSE-C measures the confidence of lip-speech synchronization, where a higher value indicates stronger alignment.
These two metrics are used as the primary synchronization metrics for human-speech videos.

We further evaluate speech quality using Word Error Rate (WER), Mel-Cepstral Distortion (MCD), and UTMOS.
WER is computed using Whisper~\cite{radford2023whisper} and reflects the intelligibility of generated speech.
MCD is computed with ESPnet~\cite{watanabe2018espnet} and measures acoustic distortion between speech features.
UTMOS~\cite{saeki2022utmos} estimates perceptual speech naturalness, where higher scores indicate better speech quality.
For video quality, we report Fréchet Video Distance (FVD) and Kernel Video Distance (KVD), both computed using an I3D backbone~\cite{carreira2017i3d}.
We also report CLIP score~\cite{radford2021CLIP} to measure text-video alignment.

\paragraph{Ambient-sound evaluation.}
For ambient-sound videos, we use DeSync as the primary temporal synchronization metric.
Following MMAudio~\cite{cheng2025MMAudio}, DeSync is computed as the average absolute temporal offset between video and audio estimated by Synchformer~\cite{iashin2024synchformer}.
A lower DeSync value indicates better temporal alignment between visual events and generated audio.

To measure video-audio semantic consistency, we report VA-IB, computed as the embedding similarity between video and audio using ImageBind~\cite{girdhar2023imagebind}.
A higher VA-IB score indicates stronger semantic correspondence between the two modalities.
For audio quality, we report Fréchet Audio Distance (FAD) using two pretrained audio backbones: PANN~\cite{kong2020panns} and PaSST~\cite{koutini22passt}.
For video quality, we report FVD and KVD with an I3D backbone~\cite{carreira2017i3d}.
In addition, we use CLAP~\cite{msclap} to evaluate text-audio alignment and CLIP~\cite{radford2021CLIP} to evaluate text-video alignment.

\paragraph{Evaluation protocol.}
All metrics follow the implementations and evaluation protocols used in prior video-audio generation works~\cite{cheng2025MMAudio,guan2025taming}.
For synchronization metrics, we primarily focus on LSE-D/LSE-C for human-speech videos and DeSync for ambient-sound videos.
Other quality and alignment metrics are reported as controls to verify that synchronization improvements do not come at the cost of speech quality, audio quality, video quality, or text-modality alignment.

\subsection{Training parameters}
\label{appendix:tab_hyperparams}

The training, inference and data preprocessing configuration of vanilla DPO and SyncDPO is summarized in Tab.~\ref{tab:hyperparams}. 
Unless otherwise specified, all experiments follow these settings to ensure a fair and consistent comparison across different variants. 

\begin{table}[h!]
\centering
\caption{Training and inference hyperparameters of DPO and SyncDPO.}
\label{tab:hyperparams}

\renewcommand{\arraystretch}{1.35}
\setlength{\tabcolsep}{6pt}

\begin{tabular}{l l}
\toprule[1.2pt]
\textbf{Parameter} & \textbf{Value} \\
\midrule

\textit{Training Configuration} \\
\midrule
Optimizer & Adam \\
Learning Rate & $5 \times 10^{-6}$ \\
Weight Decay & $1 \times 10^{-4}$ \\
Adam Betas & $(0.9, 0.95)$ \\
Adam Epsilon & $1 \times 10^{-8}$ \\
LR Scheduler & Cosine decay with linear warmup \\
Warmup Steps & 1,000 \\
EMA Decay & 0.9 \\
DPO Coefficient ($\beta$) & 0.2 \\
Curriculum Rate ($k$) & $0.01\%$ \\
LoRA Rank & 128 \\
LoRA Scaling ($\alpha$) & 64 \\
Training Precision & Mixed precision (fp16/bf16) \\
First Frame Reference & Provided \\
Framework & PyTorch \\
Hardware & 4 $\times$ NVIDIA A800 GPUs \\

\midrule
\textit{Inference Configuration} \\
\midrule
Sampling Steps & 30 \\
Video CFG Scale & 4.0 \\
Audio CFG Scale & 3.0 \\

\midrule
\textit{Data Configuration} \\
\midrule
Target Duration & $5.0$ seconds \\
Video Resolution & $992 \times 512$ \\
Video FPS & 24 \\
Audio Sample Rate & 16 kHz \\
Audio Channels & 1 (mono) \\

\bottomrule[1.2pt]
\end{tabular}
\end{table}

\section{Experiments}
\subsection{Experiments on different construction methods}
\label{appendix:exp_construction}

\begin{table}[t!]
\centering
\caption{Ablation on different negative construction strategies across multiple benchmarks. Training Data denotes the datasets used for training. The blue-highlighted metrics indicate synchronization performance. \textbf{Bold} and \underline{underline} denote the best and second-best results within each group.}
\label{tab:detailed_cons}

\renewcommand{\arraystretch}{1.35}
\setlength{\tabcolsep}{2.0pt}

\resizebox{\textwidth}{!}{
\begin{tabular}{lcl cc cc cc cc}
\toprule[1.4pt]

\multicolumn{11}{c}{\textbf{Human Speech Benchmark}} \\
\midrule[1.4pt]

\multirow{2}{*}{Benchmark} & 
\multirow{2}{*}{Data} & 
\multirow{2}{*}{Method} & 
\multicolumn{2}{c}{Lip Alignment} & 
\multicolumn{3}{c}{Speech Quality} & 
\multicolumn{3}{c}{Video Quality} \\
\cmidrule(lr){4-5}
\cmidrule(lr){6-8}
\cmidrule(lr){9-11}

 & & & \cellcolor{blue!10}LSE-D $\downarrow$ & \cellcolor{blue!10}LSE-C $\uparrow$ & WER $\downarrow$ & MCD $\downarrow$ & UTMOS $\uparrow$ & FVD $\downarrow$ & KVD $\downarrow$ & CLIP $\uparrow$ \\
\midrule[1.4pt]

\multirow{6}{*}{LRS2}
& - & No tune 
& \cellcolor{blue!10}8.15 & \cellcolor{blue!10}7.08 
& 15.40 & 19.06 & 3.34 & 250.22 & 0.43 & 0.21 \\
\cline{2-11}
& \multirow{5}{*}{LRS2}
& Shift      
& \cellcolor{blue!10}8.33 & \cellcolor{blue!10}6.57 
& 15.90 & 18.65 & 3.31 & 309.73 & 0.52 & 0.21 \\
& & Scale      
& \cellcolor{blue!10}\underline{7.96} & \cellcolor{blue!10}\textbf{7.12} 
& 15.70 & 18.88 & 3.39 & 255.11 & 0.46 & 0.21 \\
& & Mask       
& \cellcolor{blue!10}8.61 & \cellcolor{blue!10}6.60 
& 16.20 & 18.52 & 3.22 & 406.58 & 0.64 & 0.21 \\
& & Synthesize 
& \cellcolor{blue!10}8.04 & \cellcolor{blue!10}6.80 
& 18.40 & 18.95 & 3.28 & 309.44 & 0.48 & 0.21 \\
& & Replace    
& \cellcolor{blue!10}\textbf{7.92} & \cellcolor{blue!10}\underline{7.07} 
& 15.10 & 18.50 & 3.38 & 268.98 & 0.43 & 0.21 \\

\midrule[1.4pt]

\multicolumn{11}{c}{\textbf{Ambient Sound Benchmarks}} \\
\midrule[1.4pt]

\multirow{2}{*}{Benchmark} & 
\multirow{2}{*}{Data} & 
\multirow{2}{*}{Method} & 
\multicolumn{2}{c}{VA Alignment} & 
\multicolumn{2}{c}{Audio Quality} & 
\multicolumn{2}{c}{Video Quality} & 
\multicolumn{2}{c}{Text Alignment} \\
\cmidrule(lr){4-5}
\cmidrule(lr){6-7}
\cmidrule(lr){8-9}
\cmidrule(lr){10-11}

 & & & \cellcolor{blue!10}DeSync $\downarrow$ & VA-IB $\uparrow$ & $\mathrm{FAD}_{pan.}$ $\downarrow$ & $\mathrm{FAD}_{pas.}$ $\downarrow$ & FVD $\downarrow$ & KVD $\downarrow$ & CLAP $\uparrow$ & CLIP $\uparrow$ \\
\midrule[1.4pt]

\multirow{6}{*}{AVSync15}
& - & No tune & \cellcolor{blue!10}0.80 & 0.33 & 43.31 & 422.88 & 388.31 & 6.03 & 0.49 & 0.26 \\
\cline{2-11}
& \multirow{5}{*}{\shortstack[l]{AVSync \\ +GH}} 
& Shift      & \cellcolor{blue!10}\underline{0.54} & 0.33 & 41.18 & 416.51& 382.84 & 5.90 & 0.49 & 0.25 \\
& & Scale      & \cellcolor{blue!10}\textbf{0.49} & 0.33 & 42.11 & 439.85 & 372.07 & 5.95 & 0.50 & 0.25 \\
& & Replace    & \cellcolor{blue!10}\underline{0.54} & 0.33 & 41.59 & 430.91 & 388.15 & 6.21 & 0.50 & 0.26 \\
& & Synthesize & \cellcolor{blue!10}\textbf{0.49} & 0.33 & 40.20 & 419.31 & 473.27 & 6.87 & 0.50 & 0.25 \\
& & Mask       & \cellcolor{blue!10}0.93 & 0.25 & 45.59 & 446.55 & 1748.33 & 12.63 & 0.50 & 0.26 \\

\midrule

\multirow{6}{*}{\shortstack{Greatest\\Hits}}
& - & No tune & \cellcolor{blue!10}0.46 & 0.16 & 48.19 & 8.70 & 506.01 & 1.68 & 0.34 & 0.22 \\
\cline{2-11}
& \multirow{5}{*}{\shortstack[l]{AVSync \\ +GH}}
& Shift      & \cellcolor{blue!10}0.23 & 0.14 & 42.33 & 476.94 & 373.01 & 1.79 & 0.34 & 0.22 \\
& & Scale      & \cellcolor{blue!10}\textbf{0.14} & 0.13 & 39.61 & 422.46 & 351.64 & 1.62 & 0.34 & 0.22 \\
& & Replace    & \cellcolor{blue!10}0.22 & 0.14 & 39.77 & 448.20 & 367.08 & 1.74 & 0.33 & 0.22 \\
& & Synthesize & \cellcolor{blue!10}\underline{0.17} & 0.13 & 37.17 & 397.79 & 331.45 & 1.53 & 0.35 & 0.22 \\
& & Mask       & \cellcolor{blue!10}0.69 & 0.08 & 49.50 & 556.31 & 3104.81 & 4.19 & 0.32 & 0.22 \\

\midrule

\multirow{6}{*}{AV+GH}
& - & No tune & \cellcolor{blue!10}0.63 & 0.24 & 37.73 & 363.73 & 241.47 & 2.26 & 0.41 & 0.24 \\
\cline{2-11}
& \multirow{5}{*}{\shortstack[l]{AVSync \\ +GH}}
& Shift      & \cellcolor{blue!10}0.39 & 0.23 & 34.46 & 359.04 & 240.19 & 2.28 & 0.42 & 0.24 \\
& & Scale      & \cellcolor{blue!10}\textbf{0.32} & 0.23 & 33.70 & 351.59 & 229.78 & 2.22 & 0.42 & 0.24 \\
& & Replace    & \cellcolor{blue!10}\underline{0.38} & 0.24 & 33.75 & 356.70 & 248.38 & 2.36 & 0.42 & 0.24 \\
& & Synthesize & \cellcolor{blue!10}0.40 & 0.23 & 32.58 & 334.76 & 222.96 & 2.16 & 0.43 & 0.24 \\
& & Mask       & \cellcolor{blue!10}0.81 & 0.17 & 39.43 & 404.89 & 2155.14 & 7.56 & 0.41 & 0.24 \\

\bottomrule[1.4pt]
\end{tabular}
}
\end{table}

In this section, we provide a more comprehensive breakdown of our experimental findings regarding negative construction strategies. 
We adjust several hyperparameters to improve training efficiency. Specifically, we increase the learning rate to $1\times10^{-5}$ and reduce the LoRA rank to $r=32$, fix $\beta$ = 0.1, while keeping all other settings unchanged.  Under this configuration, the model is trained for 1k steps using in-domain training data. These modifications allow faster convergence without affecting the overall conclusions. Tab.~\ref{tab:detailed_cons} extends the main paper by including evaluations on the GreatestHits~\cite{owens2016greatesthits} (GH) and the combined AVSync~\cite{zhang2024avsync}+GreatestHits (AV+GH) benchmarks. These results further substantiate the effectiveness and robustness of our proposed strategies across diverse audio-visual contexts.

On the Greatest Hits dataset, the Scale method demonstrates remarkable precision. It reduces the DeSync score from $0.46$ to $0.14$, a significant improvement that highlights the model's enhanced sensitivity to micro-temporal misalignments. In the AV+GH composite benchmark, Scale and Replace consistently emerge as the top-performing strategies. Specifically, Scale achieves the best DeSync score ($0.32$), while Replace offers a balanced improvement ($0.38$) without compromising video quality or text alignment. This proves that the two methods are not overfitted to specific data distributions but rather learn a generalized sense of temporal synchronization. For other construction methods, Masking leads to a catastrophic explosion in FVD scores across all benchmarks (e.g., jumping from $241.47$ to $2155.14$ on AV+GH benchmark). While Shift and Synthesize perform well for ambient sound synchronization, they show limited gains on speech-related benchmarks LRS2~\cite{afouras2018LRS2}.

The detailed evidence reinforces the conclusion that Replace and Scale are the most effective and efficient strategies. They provide a high-fidelity signal for fine-grained temporal alignment while preserving the semantic and perceptual quality of the generated audio-visual content.

\subsection{Experiments on hyperparameter}
\phantomsection
\label{appendix:exp_hyperparams}

In this section, we conduct a systematic ablation study on the curriculum scheduling rate $k$ and the DPO coefficient $\beta$. 
The full evaluation results across all metrics are reported in Tab.~\ref{tab:ablation_k} and Tab.~\ref{tab:ablation_beta}, respectively.

For the scheduling parameter $k$, we fix $\beta=0.1$ and vary $k$ to analyze the effect of different curriculum progression rates. 
The results show that an appropriate scheduling rate is crucial for achieving optimal performance. When $k$ is too small or too large, the model exhibits suboptimal convergence. Empirically, we observe that $k=0.01$ achieves the best balance, leading to consistent improvements across both synchronization and semantic alignment metrics.

\begin{table}[h!]
\centering
\caption{Ablation on the scheduling parameter $k$. Best results are in \textbf{bold} and second-best results are \underline{underlined}.}
\label{tab:ablation_k}

\renewcommand{\arraystretch}{1.05}
\setlength{\tabcolsep}{2pt}
\resizebox{\textwidth}{!}{

\begin{tabular}{l>{\columncolor{blue!10}}c>{\columncolor{blue!10}}c  ccc>{\columncolor{blue!10}}c c cc}
\toprule[1.2pt]

& \multicolumn{5}{c}{\textbf{LRS2}} 
& \multicolumn{4}{c}{\textbf{VABench}} \\

\cmidrule(lr){2-6} \cmidrule(lr){7-10}

$k$
& \multicolumn{2}{c}{Lip Alignment}
& \multicolumn{3}{c}{Speech Quality}
& \multicolumn{2}{c}{VA Alignment}
& \multicolumn{2}{c}{Text Alignment} \\

\cmidrule(lr){2-3} \cmidrule(lr){4-6} \cmidrule(lr){7-8} \cmidrule(lr){9-10}

& \cellcolor{blue!10}LSE-D $\downarrow$ 
& \cellcolor{blue!10}LSE-C $\uparrow$
& WER $\downarrow$ & MCD $\downarrow$ & UTMOS $\uparrow$
& \cellcolor{blue!10}DeSync $\downarrow$ 
& VA-IB $\uparrow$
& CLAP $\uparrow$ & CLIP $\uparrow$ \\

\midrule[1.2pt]

0
& \underline{7.86} & 7.19 & 15.50 & 19.04 & 3.40
& \underline{0.59} & \underline{0.24} & 0.39 & 0.24 \\

0.005
& \underline{7.86} & 7.18 & 15.50 & 18.84 & 3.41
& 0.60 & \textbf{0.25} & 0.40 & 0.23 \\

0.010
& \textbf{7.83} & \textbf{7.26} & 15.30 & 18.86 & 3.38
& \textbf{0.54} & \textbf{0.25} & 0.40 & 0.23 \\

0.015
& 7.87 & \underline{7.21} & 15.20 & 18.84 & 3.38
& \underline{0.59} & 0.23 & 0.39 & 0.24 \\

\bottomrule[1.2pt]
\end{tabular}
}
\end{table}

For the study of the DPO coefficient $\beta$, we fix $k=0.01$ and vary $\beta$ to examine its influence on preference learning. 
The value of $\beta$ controls the strength of preference optimization and thus significantly affects training dynamics. A small $\beta$ may lead to insufficient preference signal, while a large $\beta$ can cause overly aggressive updates and degrade stability. Our results indicate that $\beta=0.2$ provides the best trade-off, yielding strong performance across both LRS2 and VABench benchmarks.

\begin{table*}[h!]
\centering
\caption{Ablation on the DPO coefficient $\beta$. Best results are in \textbf{bold} and second-best results are \underline{underlined}.}
\label{tab:ablation_beta}

\renewcommand{\arraystretch}{1.05}
\setlength{\tabcolsep}{2pt}
\resizebox{\textwidth}{!}{

\begin{tabular}{c>{\columncolor{blue!10}}c>{\columncolor{blue!10}}c ccc >{\columncolor{blue!10}}c c cc}
\toprule[1.2pt]

& \multicolumn{5}{c}{\textbf{LRS2}} 
& \multicolumn{4}{c}{\textbf{VABench}} \\

\cmidrule(lr){2-6} \cmidrule(lr){7-10}

$\beta$ 
& \multicolumn{2}{c}{Lip Alignment} 
& \multicolumn{3}{c}{Speech Quality} 
& \multicolumn{2}{c}{VA Alignment} 
& \multicolumn{2}{c}{Text Alignment} \\

\cmidrule(lr){2-3} \cmidrule(lr){4-6} \cmidrule(lr){7-8} \cmidrule(lr){9-10}

& \cellcolor{blue!10}LSE-D $\downarrow$ 
& \cellcolor{blue!10}LSE-C $\uparrow$
& WER $\downarrow$ & MCD $\downarrow$ & UTMOS $\uparrow$
& \cellcolor{blue!10}DeSync $\downarrow$ 
& VA-IB $\uparrow$
& CLAP $\uparrow$ & CLIP $\uparrow$ \\

\midrule[1.2pt]

0.1 
& \textbf{7.83} & \textbf{7.26} & 15.30 & 18.86 & 3.38
& 0.54 & \textbf{0.25} & 0.40 & 0.23 \\

0.2 
& \textbf{7.83} & \underline{7.18} & 15.30 & 18.90 & 3.38
& \underline{0.43} & \textbf{0.25} & 0.39 & 0.23 \\

0.3 
& \underline{7.88} & 7.11 & 16.00 & 18.90 & 3.42
& \underline{0.43} & 0.23 & 0.38 & \underline{0.23} \\

0.4 
& 7.91 & 7.00 & 15.40 & 18.90 & 3.45
& \textbf{0.41} & 0.23 & 0.38 & \underline{0.23} \\

\bottomrule[1.2pt]
\end{tabular}
}
\end{table*}

Therefore, in our main experiments, we set $k=0.01$ and $\beta=0.2$ based on the ablation results (see~\ref{tab:hyperparams}).

\section{Limitations and broader impact.}
\label{appendix:limit&impact}
We anticipate that scaling up the training data could further amplify the benefits of preference-based temporal alignment and reveal potential scaling effects, leading to improved robustness and generalization.
Besides, the applicability of the proposed method to long videos, multi-source audio scenarios, and events with weak visual cues requires further investigation.
For broader impact, this work may benefit applications such as content creation. However, enhanced generation quality may also increase the risk of misuse, such as producing more realistic manipulated or misleading multimedia content. To mitigate these risks, we advocate for responsible use of generative models, including appropriate content moderation, watermarking, and usage restrictions.




\end{document}